\documentclass[letterpaper, 10 pt, conference]{ieeeconf}  

\IEEEoverridecommandlockouts                              

\usepackage{url}
\usepackage[hidelinks]{hyperref}
\usepackage{graphicx}
\usepackage[nolist]{acronym}
\usepackage{todonotes}
\usepackage{bm}
\usepackage{booktabs}
\usepackage{tabularx}
\usepackage{threeparttable}
\usepackage{arydshln} 
\usepackage{multirow}
\usepackage{amsmath}
\usepackage{comment}

\acrodef{ai}[AI]{artificial intelligence}
\acrodef{ml}[ML]{machine learning}
\acrodef{dof}[DoF]{degree of freedom}
\acrodef{pcb}[PCB]{printed circuit board}
\acrodef{spi}[SPI]{Shadow Program Inversion}
\acrodef{artm}[ARTM]{ArtiMinds Robot Task Model}
\acrodef{kpi}[KPI]{Key Performance Indicator}
\acrodef{mutt}[MuTT]{Multimodal Trajectory Transformer}
\acrodef{ood}[OOD]{out-of-distribution}

\newcommand{\veryshortarrow}[1][3pt]{\mathrel{%
   \vcenter{\hbox{\rule[-.5\fontdimen8\textfont3]{#1}{\fontdimen8\textfont3}}}%
   \mkern-4mu\hbox{\usefont{U}{lasy}{m}{n}\symbol{41}}}}
\newcommand{\opr}[2]{{\scriptsize #2} $\veryshortarrow$ #1}

\title{\LARGE \bf
AI-based Framework for Robust Model-Based Connector Mating in Robotic Wire Harness Installation
}

\author{ 
Claudius Kienle$^{1,4,*}$, Benjamin Alt$^{2,4,*,\dagger}$, Finn Schneider$^{3,4}$, Tobias Pertlwieser$^{3}$,\\Rainer Jäkel$^{4}$ and Rania Rayyes$^{3}$
\thanks{This work was supported by the German Federal Ministry of Economic Affairs and Climate Action (grant 13IK026A) and the German Federal Ministry of Education and Research (project RobInTime, grant 01IS25002B).}%
\thanks{$^*$ Equal contribution}
\thanks{$\dagger$ Corresponding author: {\tt\small benjamin.alt@uni-bremen.de}}
\thanks{\raggedright $^{1}$IAS Lab, Computer Science Department, TU Darmstadt, Germany}%
\thanks{\raggedright $^{2}$AICOR Institute for Artificial Intelligence, University of Bremen, Germany}%
\thanks{\raggedright $^{3}$Institute for Material Handling and Logistics (IFL), Karlsruhe Institute of Technology (KIT), Karlsruhe, Germany}
\thanks{\raggedright $^{4}$ArtiMinds Robotics, Karlsruhe, Germany}
}%

\begin{document}
\bstctlcite{IEEEexample:BSTcontrol}

\maketitle
\thispagestyle{empty}
\pagestyle{empty}

\global\csname @topnum\endcsname 0  
\global\csname @botnum\endcsname 0

\begin{abstract}

Despite the widespread adoption of industrial robots in automotive assembly, wire harness installation remains a largely manual process, as it requires precise and flexible manipulation.
To address this challenge, we design a novel AI-based framework that automates cable connector mating by integrating force control with deep visuotactile learning. Our system optimizes search-and-insertion strategies using first-order optimization over a multimodal transformer architecture trained on visual, tactile, and proprioceptive data. Additionally, we design a novel automated data collection and optimization pipeline that minimizes the need for machine learning expertise. The framework optimizes robot programs that run natively on standard industrial controllers, permitting human experts to audit and certify them. Experimental validations on a center console assembly task demonstrate significant improvements in cycle times and robustness compared to conventional robot programming approaches. Videos are available under 
\url{https://claudius-kienle.github.io/AppMuTT}.

\end{abstract}

\section{INTRODUCTION}

\begin{figure}
\includegraphics[width=\linewidth]{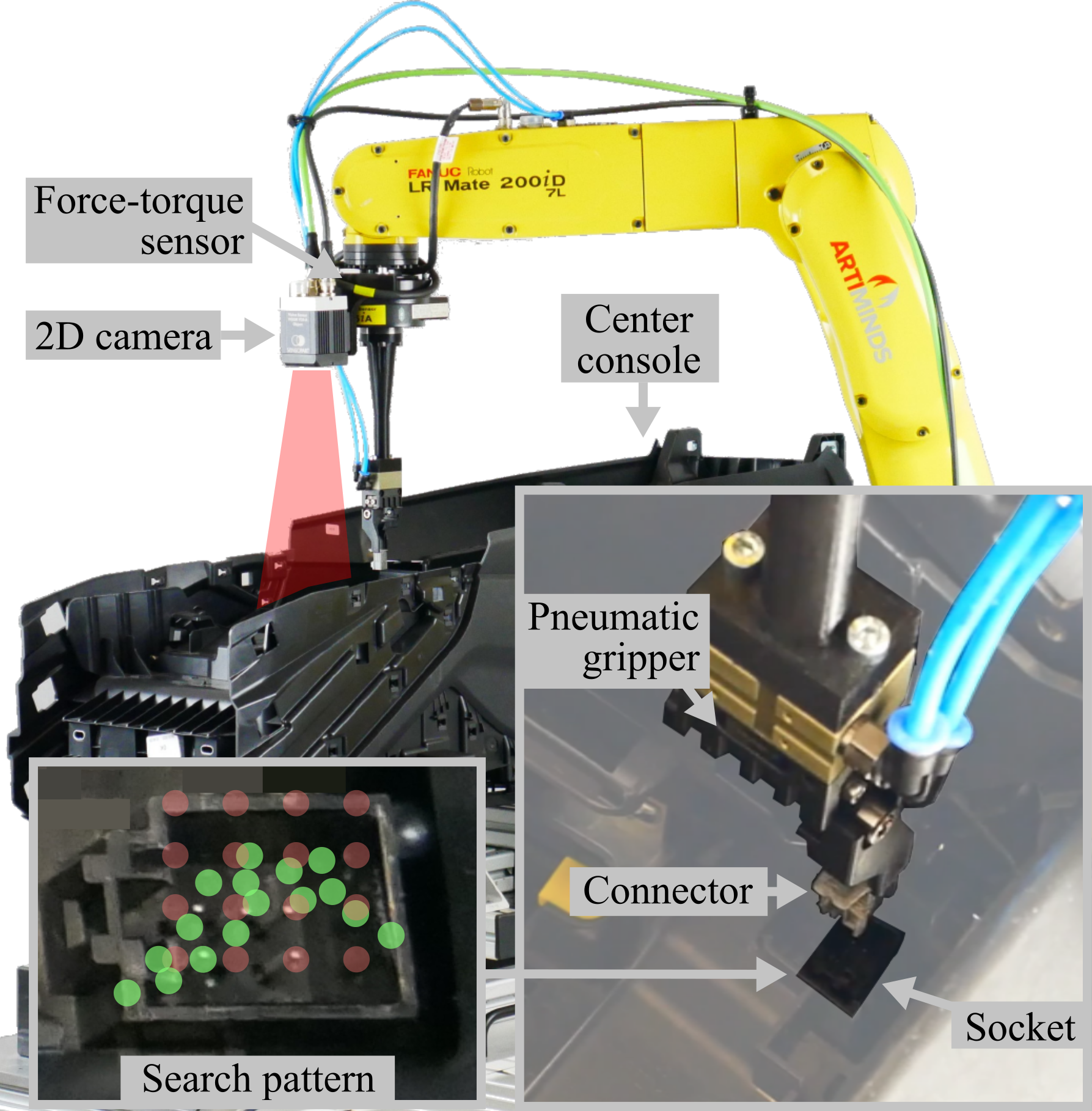}
\caption{Robotic mating of electrical connectors in a center console: Using our learned, neural model of the search-and-insertion strategy, the search pattern (bottom left, red) is optimized (green) to ensure robust installation with minimal temporal overhead.}
\label{fig:teaser_img}
\end{figure}

Industrial robots have been central to the automation of assembly tasks in the automotive industry for several decades \cite{hagele_industrial_2016}. Recent advances in sensor and actor technologies such as cost-effective force-torque sensors and 3D cameras have facilitated software- and data-driven automation, enabling the use of robots in increasingly dynamic, flexible assembly applications \cite{bartos_overview_2021}. Despite these advances, the installation of wire harnesses has remained largely confined to manual labor and expensive, specialized mechanical engineering solutions \cite{wnuk_case_2023}. 
This is largely due to the process variances induced by the flexibility of cables, the wide range of product variants, and strict process requirements with respect to cycle time and robustness \cite{trommnau_overview_2019}.
As modern automobiles incorporate more connected sensors and actuators, partly driven by the demands of autonomous driving, the number and complexity of wire harnesses are rapidly increasing. As a result, wire harness installation is becoming a greater bottleneck and cost factor in the automotive value chain \cite{trommnau_overview_2019}.

One of the final steps in wire harness installation in automobiles is the mating of the cable connectors \cite{yumbla_analysis_2019}. This installation step poses considerable challenges for robotic automation, as it requires handling a large variety of connector geometries \cite{trommnau_overview_2019} and accessing sockets located in the drivetrain, center console, dashboard, and other areas that are hard to reach for 6-\ac{dof} robot arms \cite{hermansson_automatic_2013}.
Moreover, the pre-positioning of the wire harness and installation targets is subject to stochastic variances on the order of several millimeters \cite{wnuk_case_2023}. State-of-the-art techniques for robotic connector mating use force-controlled or vision-based search and insertion, inherently facing a tradeoff between robustness and execution time.

To address this challenge, we propose a novel framework for the automated mating of wire harness cable connectors, combining advanced machine learning techniques \cite{kienle_mutt_2024} with a streamlined, automation-friendly process \cite{alt_bansai_2024}. It leverages multimodal visuotactile learning and force control to optimize the connector insertion strategy, enhancing both efficiency and robustness.
At the core of our framework is a \ac{mutt} \cite{kienle_mutt_2024}, trained on visual, tactile, and proprioceptive data to capture the stochastic characteristics of the insertion process. By applying \ac{mutt} as a predictor in a model-based optimizer \cite{alt_robot_2021}, we enable the system to adapt its search-and-insertion strategy, leading to improved cycle times and insertion success rates. 
Our AI-based framework is complemented by a fully automated pipeline for data collection, training, and optimization, minimizing the need for human intervention and making the system accessible to non-experts in AI. 

We validate our framework in real-world experiments on installing a wire harness into a center console, and evaluate its performance across five connector types with different geometries. Search optimization results in updated parameters on the robot controller, ensuring that the resulting search strategy remains interpretable, auditable, and certifiable by industry professionals.

\section{RELATED WORK}

\subsection{Robotic Wire Harness Installation}

Wire harness installation features several challenging perception and manipulation problems, such as the accurate perception of the position and state of deformation of the harness \cite{wang_systematic_2024,nguyen_revolutionizing_2024,wnuk_case_2023,wang_deep_2023}, robust grasping of the wires or connectors \cite{kamiya_learning_2024,zhang_closed-loop_2024,zhang_learning_2023}, outlaying and fixing of the harness in the chassis \cite{fogt_cable_2024,hermansson_automatic_2013,jiang_robotized_2011} and mating of the electrical connectors \cite{yumbla_analysis_2019,di_vision-force_2012,sun_robotic_2010}. This paper focuses on the latter task of connector insertion, which remains a manual process in the industrial state of the art \cite{wang_deep_2023}. Researchers have proposed several approaches for robotic connector mating, most of which use a combination of visual and force feedback. Sun et al. \cite{sun_robotic_2010} propose to measure the tilt angle and displacement of the grasped connector with two 2D cameras and use a force- and torque-controlled search and insertion strategy to mate the connector with the socket. Di et al. \cite{di_vision-force_2012} propose a hybrid force-vision control strategy for online motion control. Yumbla et al. \cite{yumbla_analysis_2019} propose a gripper design that allows grasping the cable so that the connector is aligned and firmly arrested, considerably reducing the uncertainty induced by imprecise grasping. 

Our framework improves upon previous methods for force and vision-controlled connector mating \cite{sun_robotic_2010,song2017electric} by introducing a gripper-friendly connector design (cf.~Fig.~\ref{fig:endeffector}) and an optimization approach that tailors the search strategy to the stochastic variance of the insertion process and the specific characteristics of the current environment. By learning the environment distribution with \ac{mutt} and adapting the search accordingly, our method minimizes unnecessary exploration, reducing search overhead while enhancing the robustness and efficiency of the mating process.

\subsection{Data-Driven Robot Program Optimization}

Force-controlled search and insertion strategies compensate for sources of process noise, such as elastic deformation of the socket, that are difficult to detect via purely vision-based approaches. Force-controlled search and careful moment-free insertion trade off robustness for time. Data-driven methods permit to optimize search and insertion strategies to minimize additional search overhead. While a variety of general-purpose robot program optimization approaches have been proposed \cite{akrour_local_2017,liang_optimization_2017,calandra_experimental_2014,marvel_automated_2009}, model-based parameter optimization has emerged as a promising approach for peg-in-hole assembly problems, as it avoids trial and error at runtime in favor of optimizing over a learned model of the process \cite{prakash_practical_2024}. In prior work, we introduced \ac{spi}, a first-order parameter optimizer operating on a differentiable, partially learned model of a robot program \cite{alt_robot_2021,alt_heuristic-free_2022}. Related approaches have applied optimization by gradient descent over differentiable programs to optimize motion controller parameters \cite{jatavallabhula_bayesian_2023,qiao_scalable_2020,degrave_differentiable_2019,hu_difftaichi_2019}. 

\ac{spi} has been applied to peg-in-hole electronics assembly \cite{alt_heuristic-free_2022}, but without conditioning on the current environment, limiting optimization to process variance alone. In prior work \cite{kienle_mutt_2024}, we extended \ac{spi} with \ac{mutt} to enable environment-conditioned optimization of the search strategy, demonstrating its effectiveness in a controlled laboratory peg-in-hole assembly experiment. However, in that setting, the process was comparatively robust to small-scale variance, permitting the system to infer the connector’s position primarily from visual information. As a result, the search strategy optimization was dominated by the adaptation to the current environment and the optimization for process variance was less significant.

In contrast, connector mating for wire harness installation presents a more fine-grained challenge. Sub-millimeter tolerances cause the process to be sensitive to variances below the resolution of vision sensors. In this work, we present the first application of \ac{mutt} \cite{kienle_mutt_2024} with \ac{spi} \cite{alt_heuristic-free_2022} for high-precision robot tasks, demonstrating its ability to optimize search strategies that account for both the observed environment and process variance in the real world. It enables search strategy optimization to sub-millimeter precision, ensuring more efficient and reliable mating, even in scenarios where vision data alone cannot fully capture process variations.

\section{AI-BASED FRAMEWORK FOR AUTOMATED CONNECTOR MATING}
\begin{figure*}
    \centering
    \includegraphics[width=\textwidth]{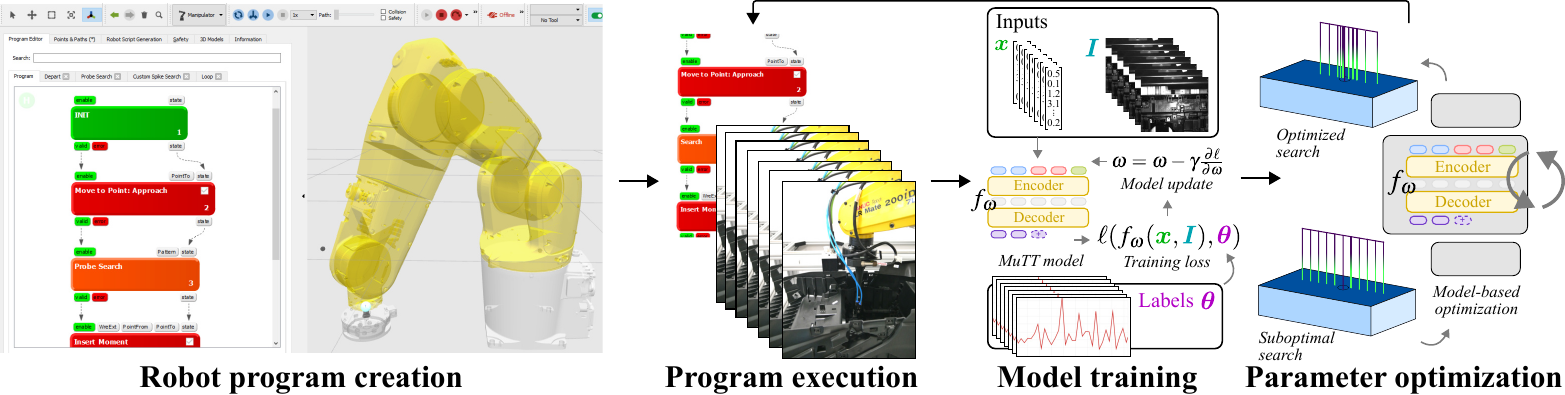}
    \caption{AI-based Framework for Visuotactile Connector Mating: A programmer creates a initial robot program using industry-standard tools (ARTM, left). During the ramp-up phase, the robot executes the program repeatedly with varying search parameterizations. The resulting dataset is used to train \ac{mutt},
    a predictive visuotactile model of the robot and environment dynamics (center). 
    MuTT serves as a sim2real predictor for the first-order optimizer SPI, which optimizes program parameters for robust and fast connector mating given the observed process variance and current environment image (right).}
    \label{fig:overview_appmutt_wide}
\end{figure*}

Our framework is shown in Fig.~\ref{fig:overview_appmutt_wide}. 
The process consists of three main stages: robot programming, learning, and optimization. 
Initially, an engineer creates a robot program using standard robot programming software (Fig.~\ref{fig:robot_program}). The program comprises a force-controlled search strategy for connector mating and is pre-parameterized by the engineer.
During ramp-up of the robot cell, the robot executes this program repeatedly.
Due to the inherent robustness of force-controlled search, the robot will successfully perform the task, albeit at the cost of search time. Details on the robot program used in our experiments, including the force-controlled search strategy, are provided in Sec.~\ref{sec:robot-program}, along with the automated data collection process for training.

Execution data including program parameters, robot trajectories, and images, is automatically collected and used to train two environment-conditioned \ac{mutt} models $m_{\omega_1}$ and $m_{\omega_2}$ for the mating process (Sec.~\ref{sec:model-learning}). 
The trained models are then used as predictors to optimize the robot program parameters end-to-end, jointly minimizing cycle time and failure probability (Sec.~\ref{sec:optimization}). 
Data from the optimized execution can be used to continuously finetune the models, realizing lifelong learning \cite{alt_heuristic-free_2022}. 
This entire process, from data collection to model training, is fully automated and configured via intuitive user interfaces \cite{alt_human-ai_2024}.
\subsection{Connector Mating with Tactile Probe Search}
\label{sec:robot-program}
Echoing related work \cite{sun_robotic_2010,alt_heuristic-free_2022}, we propose a robotic connector mating technique based on force-controlled search and insertion, but optimize the search to fit the process noise distribution using a model-based optimizer. We implement the robot program using the \ac{artm}, a commercial cross-platform robot program representation \cite{schmidt-rohr_artiminds_2013}. It represents a task as a set of parameterized skills, combined by control flow primitives (e.g. \texttt{if}, \texttt{for}, ...). Each skill accepts a set of parameters, which parameterize a motion planner and a runtime controller. 

The search-and-insertion program for connector mating is shown in Fig. \ref{fig:robot_program}. The \texttt{Probe Search} skill executes three sub-motions in a loop (cf.~Fig.~\ref{fig:overview_appmutt_wide}, right): A linear motion to the next probing position; a force-controlled linear motion along the local Z axis, which stops on contact with the surface; and a linear depart motion along the local Z axis. When the contact motion fails (the relative motion defined by parameters \texttt{PointFrom} and \texttt{PointTo} is executed without exceeding the defined contact forces), the search ends prematurely and returns success (\texttt{valid} is set to \texttt{True}). This and other error handling logic are omitted in Fig. \ref{fig:robot_program} for brevity.

\subsection{Learning an Environment-Aware Forward Model}
\label{sec:model-learning}

\begin{figure}
    \centering
    \includegraphics[width=\linewidth]{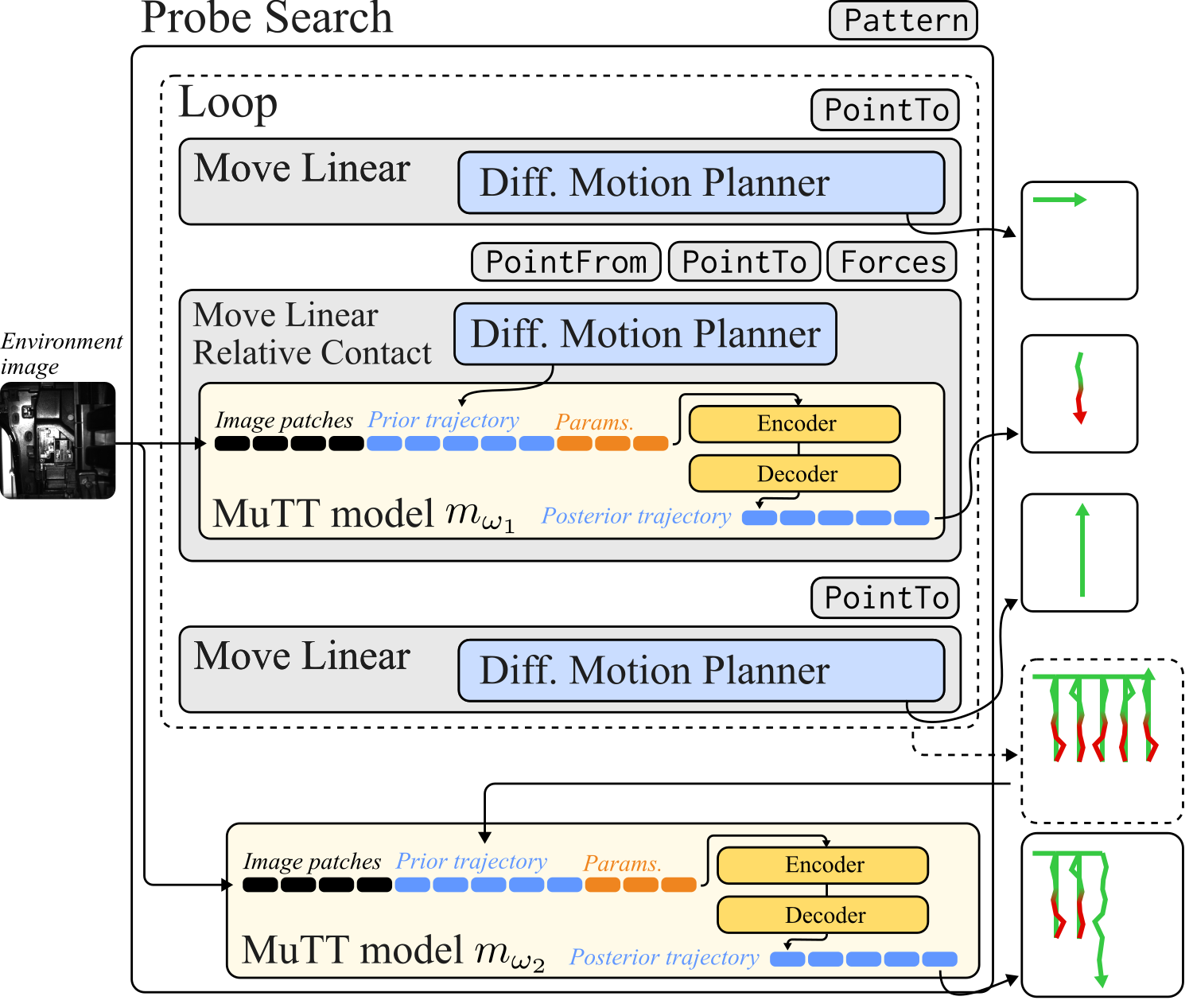}
    \caption{Shadow program architecture for probe search. Differentiable motion planners (blue) \cite{alt_shadow_2024} and multimodal Transformers (yellow) \cite{kienle_mutt_2024} are combined in a differentiable computation graph. Forward evaluation (top to bottom) yields a prediction of the expected trajectory $\bm{\bar{\theta}}$ given the program parameters (\texttt{Pattern}, \texttt{PointTo} etc.).}
    \label{fig:mutt_architecture}
\end{figure}

A probe search is pre-parameterized by an engineering expert for a particular connector and searches on a grid, which incurs considerable search overhead. 
We seek to adapt the search to optimally fit the current real-world noise distribution. Moreover, we incorporate visual information into the optimization, so that the search pattern is optimized \textit{online} given an image of the environment. To that end, we implement \ac{spi} \cite{alt_robot_2021}, a first-order optimizer over a learned, differentiable model (``shadow progam'', $\bar{P}_{\omega}$) of the \ac{artm} robot program $P$. The model architecture is shown in Fig. \ref{fig:mutt_architecture}. The shadow program combines Cartesian \cite{alt_robot_2021} and joint-space \cite{alt_shadow_2024} differentiable motion planners with neural networks and predicts the expected robot trajectory $\bm{\bar{\theta}}$, including joint configurations and end-effector forces and torques, given a set of task parameters $\bm{x}$ (such as \texttt{PointTo}, the starting point of a search), and an image $\bm{I}$ of the environment. The weights $\omega$ of the neural networks in $\bar{P}_{\omega}$ are learned from observation data collected by executing $P$ in the real-world environment, subject to small variations in the task parameters $\bm{x}$ and simulated process variance. For details on the model training and data collection procedure, we refer to the literature \cite{kienle_mutt_2024,alt_bansai_2024,alt_heuristic-free_2022}.

Probe search can involve very (temporally) long trajectories, as many probes might be required to find the socket. For predictive models, the difficulty of the learning problem grows significantly with the length of the predicted trajectory. For this reason, we propose to split the learning problem into two separate stages, as illustrated in \cite{kienle_mutt_2024}. First, the `inner' model for individual force-controlled probing (model $m_{\omega_1}$ in Fig. \ref{fig:mutt_architecture}) is trained on a dataset comprising all probe motions across all search skill executions in the dataset. Then, the `outer' model $m_{\omega_2}$ is trained to learn the search-level semantics, i.e. to end the trajectory once the predicted end-effector pose drops into the hole, and to set the predicted success flag to 1 if the hole was found.

\subsection{Optimizing for Robust and Fast Connector Mating}
\label{sec:optimization}

Once the predictive models $m_{\omega_1}$ and $m_{\omega_2}$ have been learned, the task parameters $\bm{x}$, comprising the target \texttt{PointTo} of the approach motion and the search pattern \texttt{Pattern} of the probe search, are optimized via gradient descent over the shadow program $\bar{P}_{\omega}$. 
With $\omega$ frozen, a forward pass is performed to obtain the predicted trajectory $\bm{\bar{\theta}}$, which is evaluated on the task-specific objective function $\Phi$ to compute the task loss. This loss is backpropagated through $\bar{P}_{\omega}$ to iteratively update $\bm{x}$, minimizing $\Phi$.
For the purpose of connector mating, $\Phi$ is the weighted sum between task success $\Phi_s$ and cycle time $\Phi_c$:
\begin{equation}
    \Phi(\bm{\bar{\theta}}) = {w_s}{\Phi_s} + {w_c}{\Phi_c} 
\end{equation}
For implementation details of the task objectives $\Phi_s$ and $\Phi_c$, we refer to prior work \cite{kienle_mutt_2024}. For the use case at hand, gradient-based optimization of task parameters $\bm{x}$ yields an optimal approach pose and search pattern to maximize the probability of successful mating with minimal temporal overhead. At runtime, the optimizer is provided with an image of the inside of the center console, the current task parameters $\bm{x}$ and the initial robot state. After optimization, the robot program $P$ is executed with the optimized parameters. Note that we do not directly `play back' the last (optimal) expected trajectory $\bm{\bar{\theta}}$ on the robot, but instead execute $P$ with the optimized parameters. This offers an additional layer of safety and reliability: By reparameterizing an industrial robot program rather than directly driving the robot, we can take advantage of existing native safety functions of the robot and the robustness of an industrial force controller.

\section{EXPERIMENTS}

\begin{figure}[tb]
    \centering
    \begin{minipage}[t]{.59\linewidth}
        \includegraphics[width=1.06\linewidth]{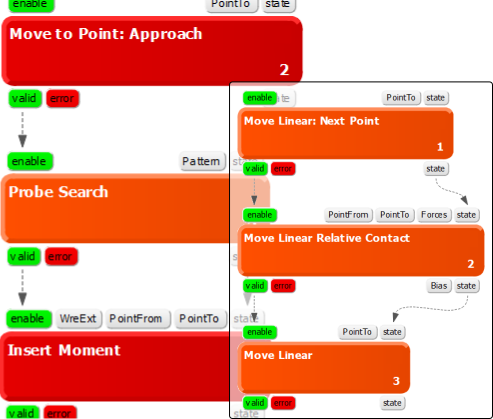}
        \caption{Extract of the \ac{artm} robot program for force-controlled search and insertion. \texttt{Probe Search} internally loops over approach, contact and depart motions for each point in the search pattern.}
        \label{fig:robot_program}
    \end{minipage}%
    \hfill%
    \begin{minipage}[t]{.35\linewidth}
        \includegraphics[width=\linewidth]{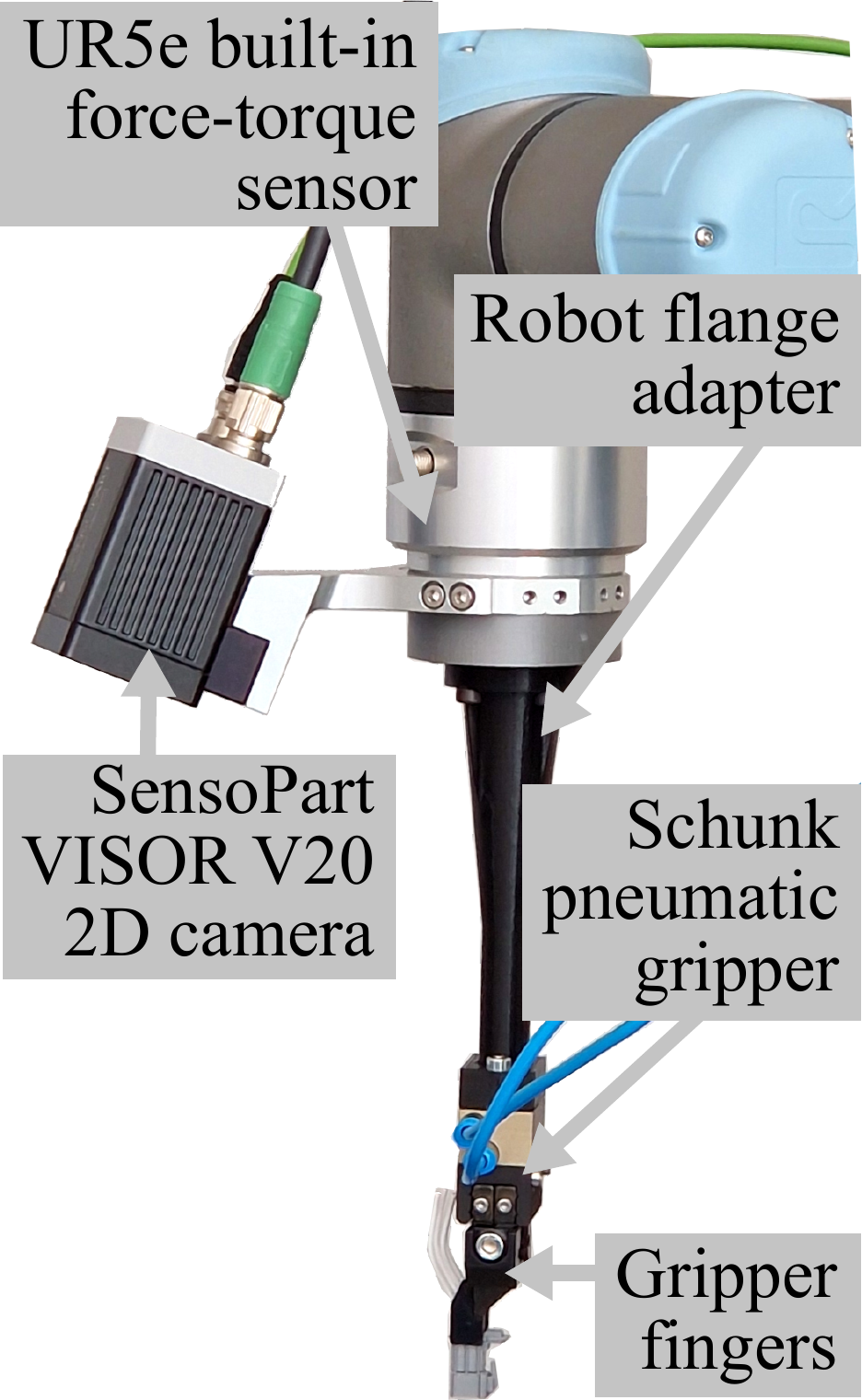}
        \caption{End-effector assembly for force-controlled connector mating. }
        \label{fig:endeffector}
    \end{minipage}
\end{figure}

We evaluate our framework on a connector mating task that reflects a robotic wire harness installation process for the center console of an automobile. The experimental setup is shown in Fig. \ref{fig:teaser_img}. 
A UR5e robot arm is equipped with a flange-mounted Sensopart VISOR V20 2D camera and a Schunk MPG-plus 32 pneumatic gripper. Fig. \ref{fig:endeffector} shows the end-effector assembly in greater detail. We consider the mating of five industrial connectors (cf.~Table \ref{tab:data_distribution}) with five different sockets in the center console of a current-generation Audi automobile. The socket for connector D is oriented with its opening to the top, the sockets for connectors A, C, and E are rotated roughly 20 degrees, and the socket for connector B is rotated by 90 degrees, opening to the side.
To facilitate robotic handling, the standard connector geometry has been adapted to include gripping ribs for form-fitting grasping with custom aluminum gripper fingers. The robot program is shown in Fig. \ref{fig:robot_program}. For each plug, the robot program is executed 4.000 times to collect training data. To approximate real-world process noise, the center console is mounted on a linear axis. At each execution, the position of the center console is perturbed by a uniformly sampled offset in the range of +/- 2 mm (+/- 1.5 mm for connector B due to space constraints). The parameters of the robot program for each execution are perturbed uniformly within a predefined domain, consisting of the approach position (+/- 2 mm), the probe pattern (each probe is uniformly sampled within 2 mm around the approach pose), as well as velocities, accelerations and contact force thresholds. End-effector motions and forces are recorded and stored alongside a 2D image of the environment, yielding a diverse dataset for each connector type (cf. Table \ref{tab:data_distribution}). Since $m_{\omega_2}$ builds upon the predictions of $m_{\omega_1}$, we train $m_{\omega_2}$ on a dataset in which the relevant parts of the ground-truth robot trajectory have been replaced by the predictions of $m_{\omega_1}$.

\begin{table}
    \centering
    \caption{Distribution of Training and Test Data}
    \label{tab:data_distribution}
    \footnotesize
    \addtolength{\tabcolsep}{-0.16em}
\newcommand{\pr}[2]{#1 {\scriptsize (#2)}}

\begin{threeparttable}
    \begin{tabularx}{\linewidth}{>{\raggedright}X rrrrr}
    & \includegraphics[width=.1\linewidth]{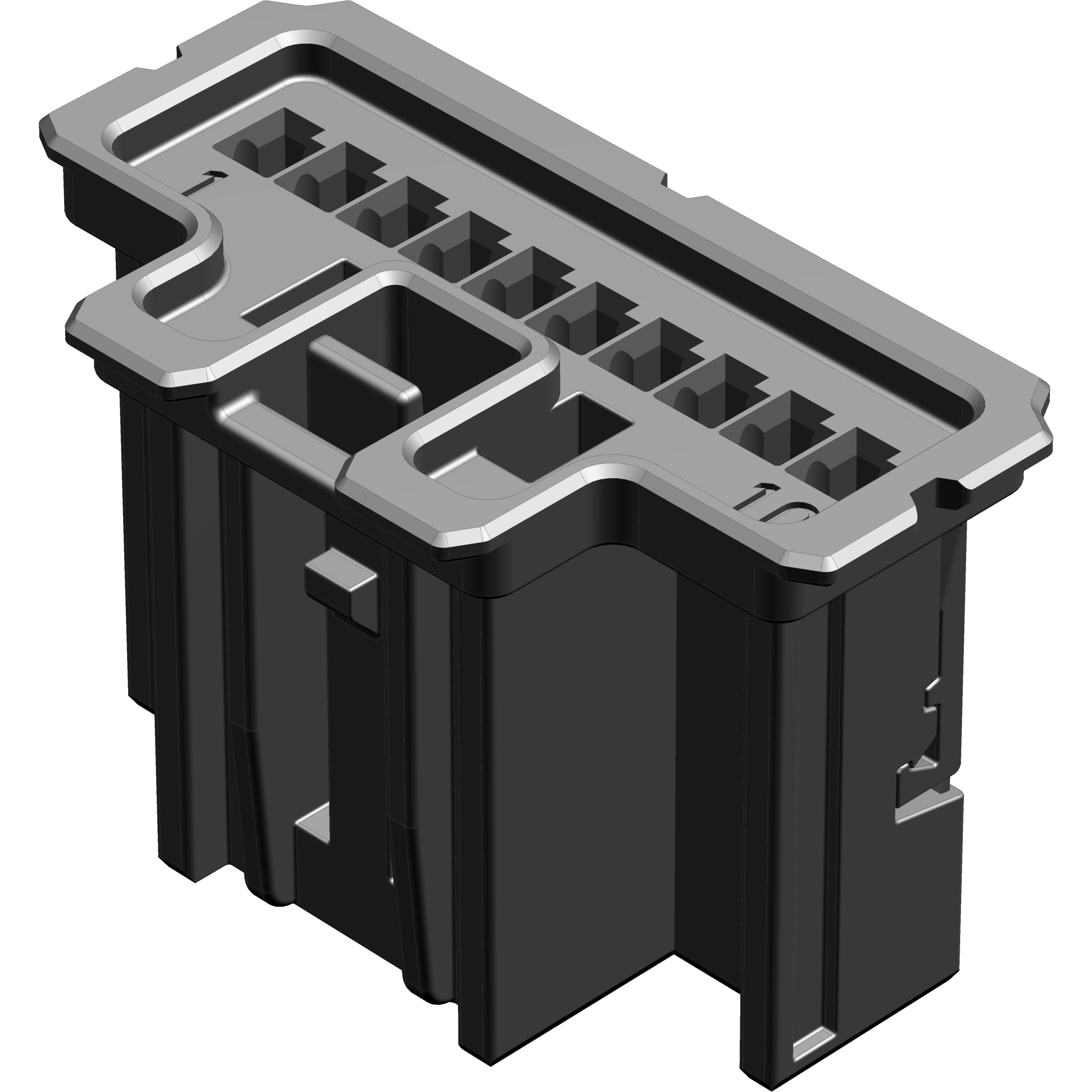} & \includegraphics[width=.1\linewidth]{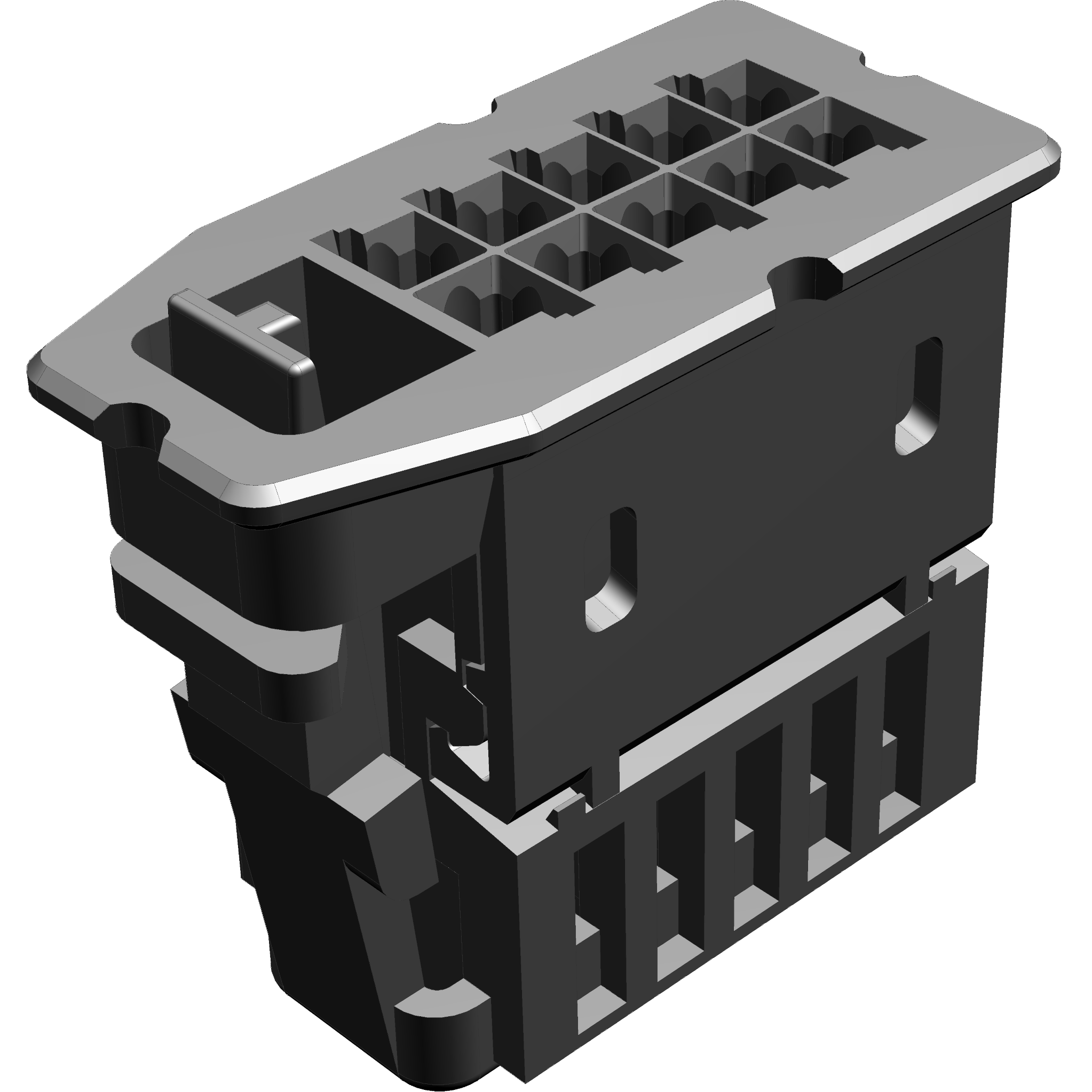} & \includegraphics[width=.1\linewidth]{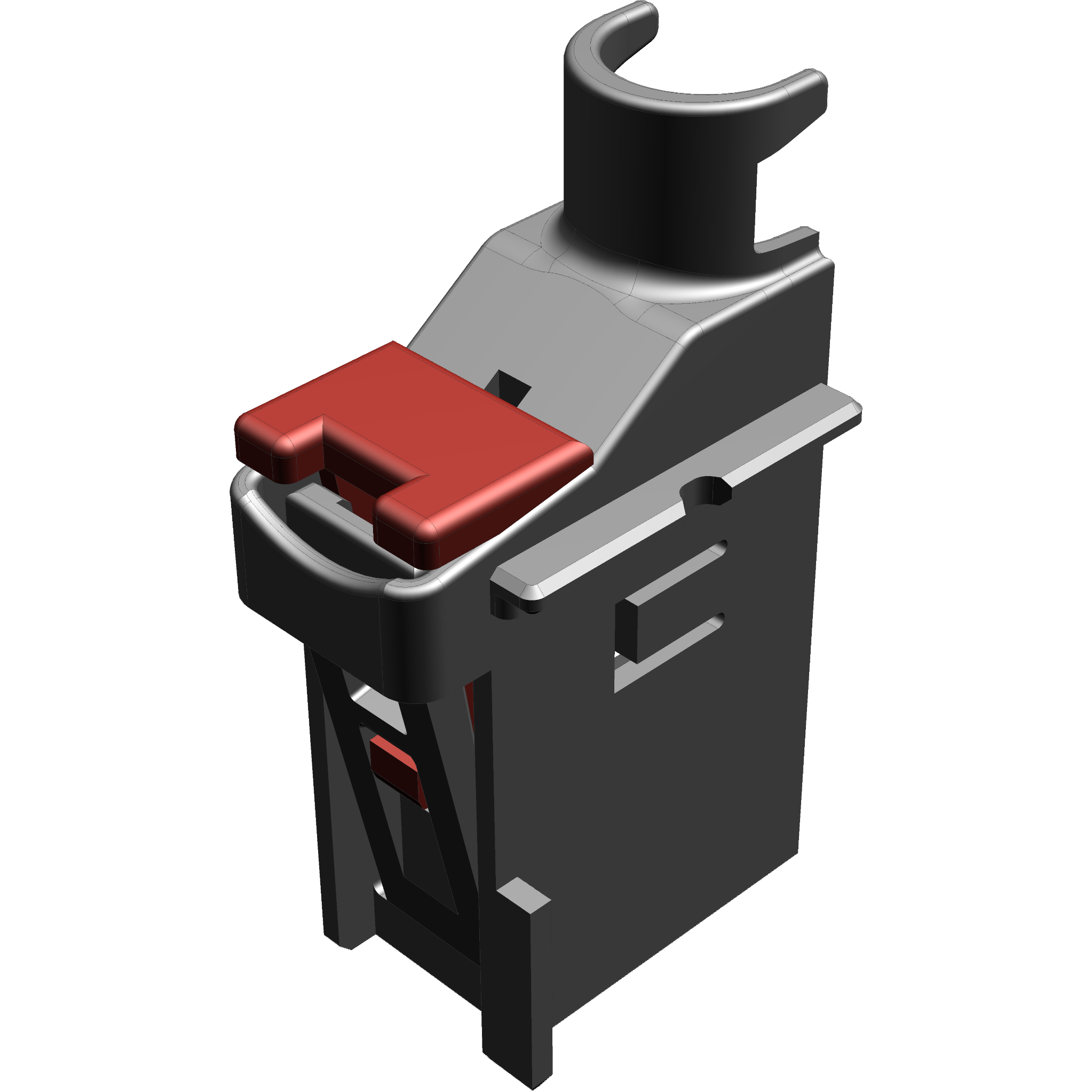} & \includegraphics[width=.1\linewidth]{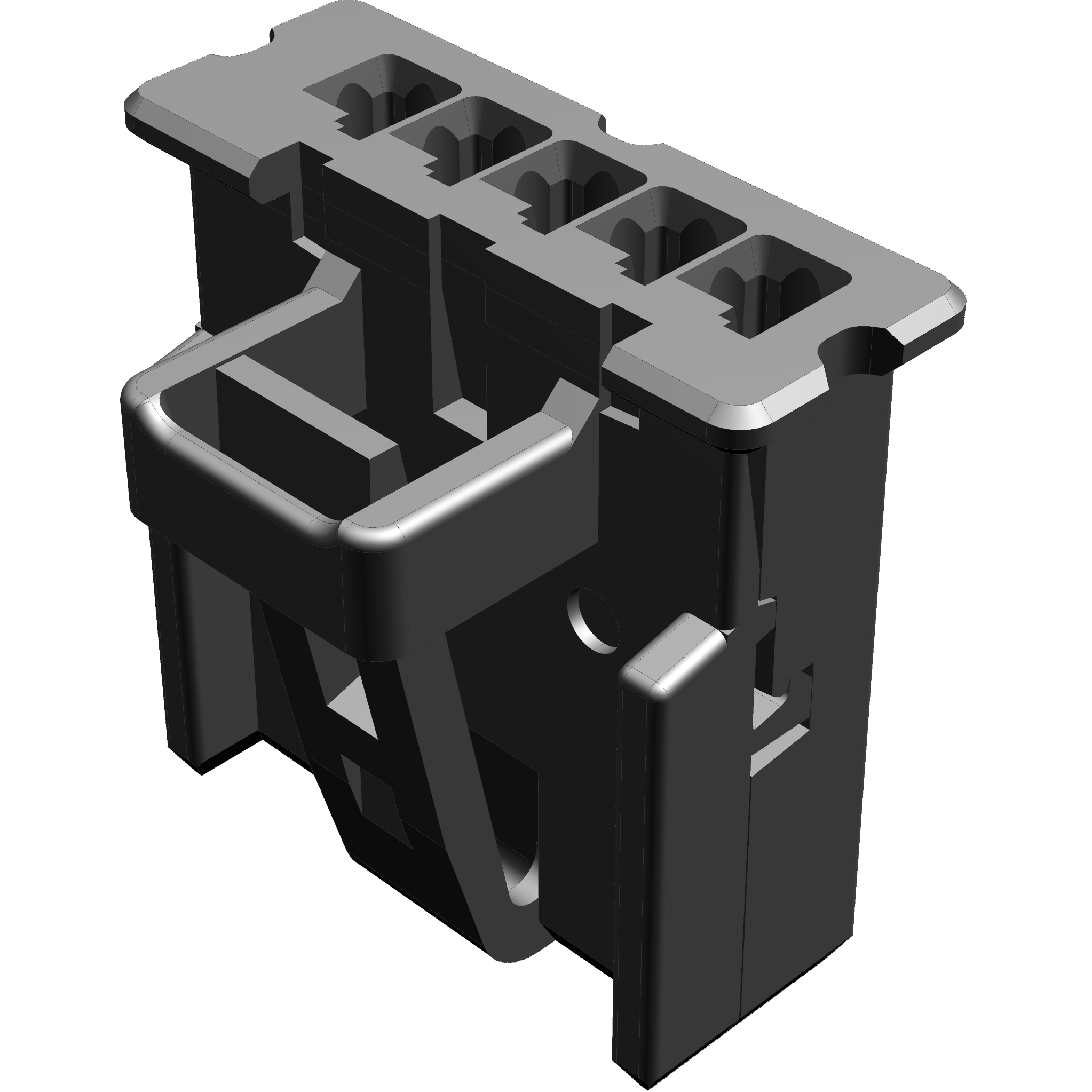} & \includegraphics[width=.1\linewidth]{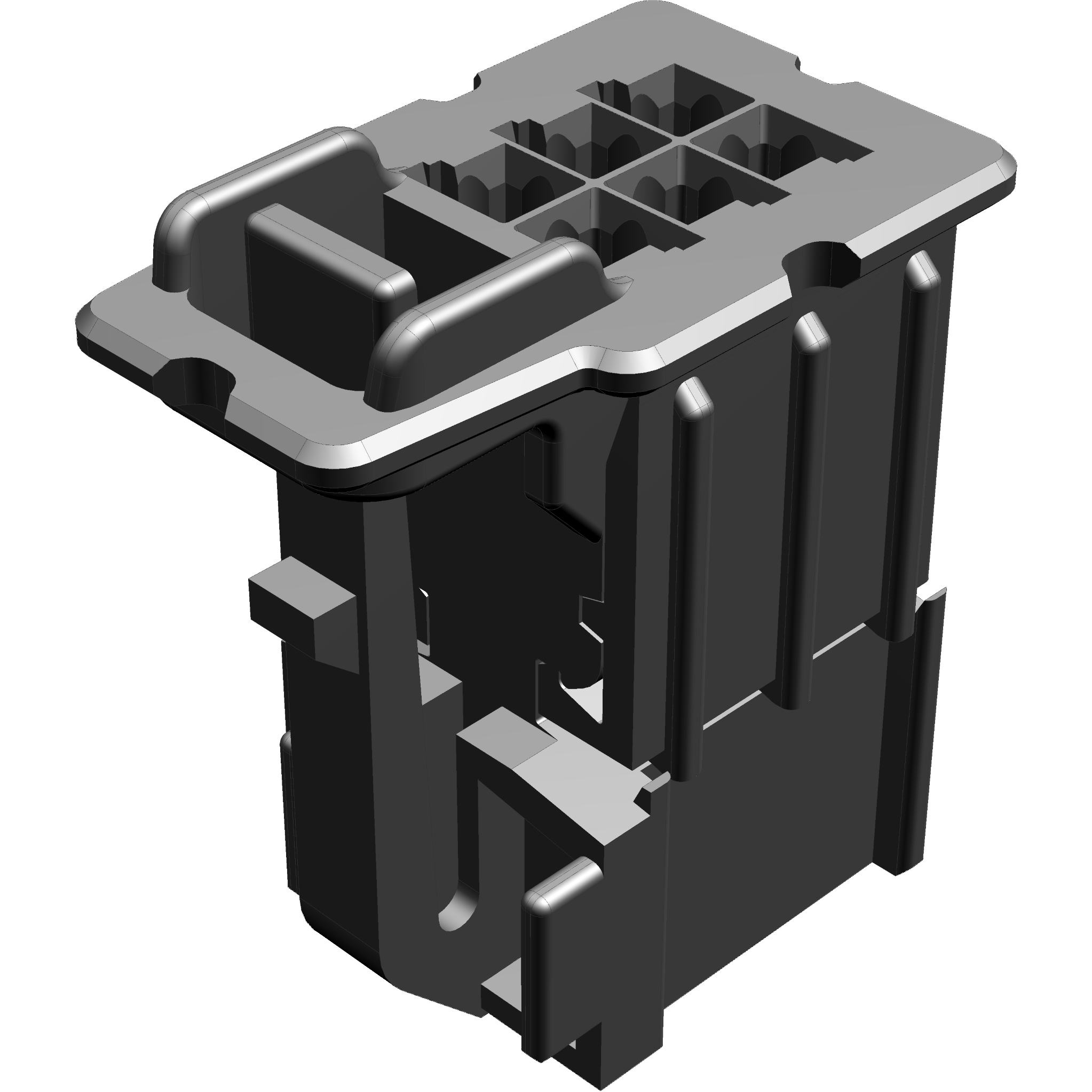} \\
        \toprule
        \textbf{Metric}     & \textbf{A}        & \textbf{B}        & \textbf{C}    & \textbf{D}    & \textbf{E}    \\\midrule
        \# train            & 3,973             & 4,000             & 3,997         & 3,996         & 3,999         \\
        \# test             & 199               & 200               & 200           & 200           & 200           \\
        \pr{Succ.}{fail}    & \pr{161}{38}      & \pr{104}{96}      & \pr{123}{77}  & \pr{172}{28}  & \pr{134}{66}  \\
        \pr{$\mu_L$}{$\sigma_L$} & \pr{201}{160} & \pr{224}{127}    & \pr{173}{131} & \pr{137}{124} & \pr{222}{151} \\
        \pr{$\mu_F$}{$\sigma_F$} & \pr{0.9}{0.8} & \pr{1.3}{1.4}    & \pr{1.2}{1.1} & \pr{1.0}{0.6} & \pr{1.1}{1.1} \\\bottomrule
    \end{tabularx}
    \begin{tablenotes}
            \centering
            \footnotesize
            \item Dataset metrics: number of training/test samples (\textit{\# train/test}), successful/failed trials (\textit{Succ}/\textit{Fail}), and mean ($\mu$) and std. dev. ($\sigma$) of trajectory length ($L$, in points) and end-effector Z-force ($F$, in N).
    \end{tablenotes}
\end{threeparttable}
\end{table}

\section{RESULTS}

\begin{table}
    \centering
    \caption{Shadow Program Prediction Results}
    \label{tab:prediction_results_composite}
    \footnotesize
    \addtolength{\tabcolsep}{-0.16em}
\newcommand{\pr}[2]{#1 {\scriptsize (#2)}}

\begin{threeparttable}
    \begin{tabularx}{\linewidth}{>{\centering\arraybackslash}m{0cm} >{\raggedright}X rrrrr}
        \toprule
         & \textbf{Metric}             & \textbf{A} & \textbf{B} & \textbf{C} & \textbf{D} & \textbf{E} \\ \midrule
        \multirow{3}{*}{%
            \rotatebox{90}{%
        \parbox{1cm}{\centering \scriptsize \textbf{Success}}%
            }%
        }
            & F1        & 99.4    & 94.2   & 98.4    & 98.6    & 98.5         \\
            & True neg. & 38/38   & 91/96  & 76/77   & 23/28  & 63/66        \\
            & True pos. & 159/161 & 97/104 & 120/123 & 172/172 & 133/134      \\
        \midrule
        \multirow{1}{*}{%
            \rotatebox{90}{%
        \parbox{1cm}{\centering \scriptsize \textbf{Traj.}}%
            }%
        }
         & $\text{MAE}_{L}$ & 21.77 & 38.47 & 13.58 & 25.01 & 17.92 \\
         & $\text{MAE}_{P}$ & 0.76  & 0.71  & 0.43  & 0.71  & 0.75 \\
         & $\text{MAE}_{F}$ & 0.98  & 1.63  & 1.09  & 0.92  & 1.03 \\
        \bottomrule
    \end{tabularx}
    \begin{tablenotes}
            \centering
            \footnotesize
            \item Evaluation of predicted task success, trajectory length ($L$, points), Cartesian position ($P$, mm), and end-effector Z-force ($F$, N).
    \end{tablenotes}
\end{threeparttable}
\end{table}

\subsection{Prediction Accuracy}
The shadow program is trained in a supervised manner for 100 epochs using an initial learning rate of $5\text{x}10^{-5}$ with linear decay. Table~\ref{tab:prediction_results_composite} presents the prediction performance of \ac{mutt} across the five connectors.
\ac{mutt} accurately predicts the expected skill execution for all five connectors, achieving an mean absolute error from the real trajectory (MAE) of the end-effector position of less than 1 mm and within 1 N of the end-effector force, which is on the order of FT sensor noise. Furthermore, \ac{mutt} predicts the expected search success with an F1 score exceeding $99~\%$ and demonstrates robust learning of the minority class (search failure) in the imbalanced datasets. The predicted trajectory length $L$ deviates by 20 points from the real world trajectory on average, corresponding to a deviation by one second from ground-truth cycle times. Overall prediction accuracies are sufficient for use in model-based optimization (Sec.~\ref{sec:optimization-results}).

\subsection{Program Optimization}
\label{sec:optimization-results}

The trained \ac{mutt} models are employed as forward models in the model-based optimizer SPI \cite{alt_robot_2021} to tune the robot program parameters for the current environment and noise distribution. We evaluated the optimization results for all five connectors on 100 unseen environments and program parameterizations (cf. Table \ref{tab:optimization_results}). These 100 samples where generated from the same distribution as the training data. The robot programs before optimization can be considered a baseline, as they perform purely tactile search. 
\begin{table}
    \centering    \caption{Optimization Results of \ac{spi} + \ac{mutt} on 100 Unseen Runs}
    \label{tab:optimization_results}
    \footnotesize
    \setlength{\tabcolsep}{5pt}
\begin{threeparttable}
\begin{tabularx}{\linewidth}{>{\raggedright}X rrrrr}
    \toprule
    \textbf{Metric} & \textbf{A} & \textbf{B} & \textbf{C} & \textbf{D} & \textbf{E} \\
    \midrule
    CT           & \opr{11}{17} & \opr{11}{18} & \opr{10}{17} & \opr{9}{13} & \opr{10}{15} \\
    \# probes & \opr{4.9}{10.2}  & \opr{5.7}{13.8} & \opr{3.0}{11.7}& \opr{3.2}{6.8}     & \opr{4.5}{9.7}      \\
    SR           & \opr{94}{75} & \opr{88}{49} & \opr{95}{50}& \opr{98}{93} & \opr{90}{72}   \\
    \bottomrule
\end{tabularx}
\begin{tablenotes}
        \centering
        \footnotesize
        \item Cycle time (CT) in seconds and success rate (SR) in \% of tasks before and after optimization (\opr{after}{before})
\end{tablenotes}
\end{threeparttable}

\end{table}
Our framework was able to optimize the programs for significantly faster and more robust insertion across all connectors. The different unoptimized search durations and success probabilities for the five connectors can be attributed mainly to the different connector and socket geometries. Connector A is comparatively larger than the other connectors and there is more play for the connector in its socket. This makes the search significantly faster, since it allows for a greater tolerance in alignment, resulting in an initial success rate of 75 \% with 10 probes fewer than the other connectors. In contrast, connectors B and C have tighter tolerances, leading to longer search durations. The probing motion for A was larger due to highly uneven surfaces around the connector. The slightly lower success rate for connector B can again be attributed to the tighter fit compared to the other connectors. Moreover, its socket is rotated by 90 degrees, requiring lateral insertion in a tightly constrained environment, which might contribute to the slightly reduced success rate of 88 \% compared to the 95 \% success rate of the other connectors.

\begin{figure}
    \centering
    \includegraphics[width=\linewidth]{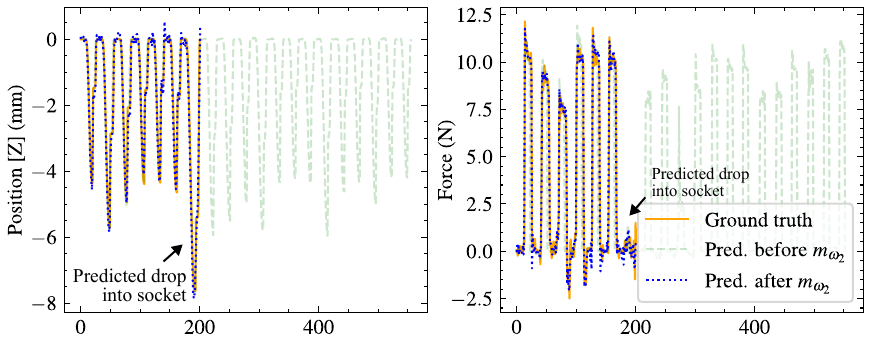}
    \caption{Predicted and ground-truth position and force trajectories for probe search (connector E). The probe-level model (green) accurately predicts the timing and force profile of individual probes.  The search-level model (blue) accurately predicts the expected time and force when the socket is found.}
    \label{fig:predicted_spikes}
\end{figure}

\subsection{Ablation Study}

Table \ref{tab:ablation} shows the results of an ablation study, in which we evaluate the performance of our framework for different sizes of the training dataset. The framework achieves optimal performance for a training dataset size of 2000, with strongly diminishing returns for dataset sizes larger than 1000. The study suggests that, while requiring sizable high-quality real-world datasets, the data collection overhead is manageable in the context of repetitive industrial assembly tasks, with the collection of 1000 training samples taking roughly three hours on our experiment setup.

\begin{table}
    \centering
    \caption{Ablation Study: Dataset Size}
    \label{tab:ablation}
    \footnotesize
    \newcommand{\pr}[2]{#1/#2}

\begin{threeparttable}
    \begin{tabularx}{\linewidth}{>{\centering\arraybackslash}m{0cm} >{\raggedright}X rrrr}
        \toprule
         & \textbf{Metric}             & 200 & 1,000 & 2,000 & 4,000  \\ \midrule
        \multirow{3}{*}{%
            \rotatebox{90}{%
        \parbox{1cm}{\centering \scriptsize \textbf{Success}}%
            }
        }
            & F1        & \underline{98.4} & 97.3 & 96.2 & \underline{98.4}       \\
            & True neg. & \pr{5}{7}    & \pr{\underline{6}}{7}    &  \pr{5}{7}   &  \pr{\underline{6}}{7}         \\
            & True pos. & \pr{\underline{92}}{93}   & \pr{89}{93}   &  \pr{88}{93}  &  \pr{91}{93}        \\
        \midrule
        \multirow{1}{*}{%
            \rotatebox{90}{%
        \parbox{1cm}{\centering \scriptsize \textbf{Traj.}}%
            }%
        }
         & $\text{MAE}_{L}$ & 1158.65 & 48.12 & 52.23 & \underline{38.17} \\
         & $\text{MAE}_{P}$ & 0.68    & 0.77  & \underline{0.59}  &  0.83 \\
         & $\text{MAE}_{F}$ & 2.48    & 0.96  & \underline{0.95}  &  0.96 \\
        \midrule
        \multirow{1}{*}{%
            \rotatebox{90}{%
        \parbox{1cm}{\centering \scriptsize \textbf{Optim.}}%
            }%
        }
         & CT           & \opr{13.4}{12.6} & \opr{9.21}{12.6} & \opr{\underline{8.6}}{12.6} & \opr{9.0}{12.6}  \\
         & \# probes & \opr{7.6}{6.8}         & \opr{3.5}{6.8}        & \opr{\underline{2.8}}{6.8}        & \opr{3.2}{6.8}  \\
         & SR           & \opr{80}{93}       & \opr{95}{93}      & \opr{\underline{98}}{93}      & \opr{\underline{98}}{93} \\
        \bottomrule
    \end{tabularx}
    \begin{tablenotes}
            \centering
            \footnotesize
            \item See Table \ref{tab:prediction_results_composite} and \ref{tab:optimization_results} for notation.
    \end{tablenotes}
\end{threeparttable}
\end{table}

\section{DISCUSSION}

Our results show that the proposed visuotactile, optimization-based search approach significantly outperforms traditional tactile-only search by effectively balancing robustness and cycle time. Moreover, we find that our approach exhibits strong robustness to illumination changes. This robustness may stem from our training data being collected over multiple days, naturally capturing illumination variations. Additionally, we have previously shown that this approach integrates with lifelong learning, enabling continuous re-optimization of robot programs \cite{alt_heuristic-free_2022} and adapting to nonstationary process noise, such as gradual drift.

In contrast to model-free self-supervised or reinforcement learning approaches, where robots learn connector mating by exploring a range of parameterized insertions across different environments, our framework avoids exploration, making training safe for industrial use. It also reduces the need for extensive manual parameter tuning to ensure robustness against process noise, streamlining industrial deployment.

Despite its advantages, our approach requires an initial training phase, introducing some overhead during ramp-up. Specifically, users must define the domain from which program parameters are sampled, leading to temporarily reduced efficiency in the early stages of deployment. This overhead largely depends on the amount of training data required. It is open to future work to explore how to further minimize the necessary dataset size, and investigate techniques to further improve \ac{ood} generalization.
Another limitation is the computational overhead associated with real-time optimization. Each execution cycle currently requires approximately 1.5 seconds, with each iteration consisting of 20 probe loops, 21 \ac{mutt} inferences, and 60 calls to the motion planner. While this performance is sufficient for many applications, further algorithmic improvements could enhance efficiency. Given that optimization time scales with computational resources, leveraging specialized hardware may yield further performance improvements.

\section{CONCLUSION}

We proposed a novel framework for robust connector mating in automated wire harness assembly. By learning an environment-conditioned forward model of the assembly task, we demonstrated model-based program parameter optimization for a more robust and efficient mating process. Our approach, validated on five different connectors, effectively accounts for process noise and adapts the program to maximize robustness while ensuring optimal search strategies based on the current environment.

The optimization quality largely depends on the prediction accuracy of the learned forward model. Future work should further investigate prediction accuracy across different settings, enhance \ac{ood} generalization, and improve few-shot learning for related tasks. The inference speed of the forward model is a key factor influencing optimization duration. Enhancing its efficiency enables real-time program optimization, bringing this approach closer to online deployment.

\bibliographystyle{IEEEtran}
\bibliography{bibliography_config,bibliography_amr,bibliography_custom}

\end{document}